\documentclass[sigconf,nonacm]{acmart}

\usepackage{soul}
\usepackage{hyperref}
\usepackage{enumitem}
\usepackage{listings}
\usepackage{cleveref}
\usepackage{booktabs} 
\usepackage{xspace}
\usepackage{microtype}
\usepackage{subcaption}
\usepackage{balance}
\usepackage{mathtools}
\usepackage{scalerel}
\usepackage{tcolorbox}
\usepackage[section]{placeins}
\usepackage{float}
\usepackage{lipsum}
\usepackage{fancyvrb}
\usepackage{algorithm}
\usepackage[noend]{algpseudocode}
\usepackage[detect-all]{siunitx}
\usepackage{mathtools}
\usepackage[english]{babel}
\usepackage{amsthm}

\DeclarePairedDelimiter\floor{\lfloor}{\rfloor}

\theoremstyle{definition}
\newtheorem{definition}{Definition}[section]

\newcommand{\vispim}{DIVAN}

\makeatletter

\AtBeginDocument{%
  }

\setcopyright{acmlicensed}
\copyrightyear{2018}
\acmYear{2018}
\acmDOI{XXXXXXX.XXXXXXX}

\acmConference[SIGMOD 2025]{}{June 22--27,
  2025}{Woodstock, NY}
\acmISBN{978-1-4503-XXXX-X/18/06}

\newcommand{\reminder}[1]{ [[[ \marginpar{\mbox{$<==$}} #1 ]]] }
\newcommand{\eat}[1]{}
\newcommand{\shift}{\mbox{\em shift}}

\begin{document}

\title{A Data Aggregation Visualization System supported by Processing-in-Memory}

\author{Junyoung Kim}
\affiliation{%
  \institution{Columbia University}
  \city{New York}
  \country{USA}
}
\email{junyoung2@cs.columbia.edu}

\author{Madhulika Balakumar}
\affiliation{%
  \institution{Columbia University}
  \city{New York}
  \country{USA}
}
\email{mb5144@columbia.edu}

\author{Kenneth Ross}
\affiliation{
  \institution{Columbia University}
  \city{New York}
  \country{USA}
}
\email{kar@cs.columbia.edu}

\begin{abstract}
Data visualization of aggregation queries is one of the most common ways of doing data exploration and data science as it can help identify correlations and patterns in the data. We propose \vispim{}, a system that automatically normalizes the one-dimensional axes by frequency to generate large numbers of two-dimensional visualizations. \vispim{} normalizes the input data via binning to allocate more pixels to data values that appear more frequently in the dataset.
\vispim{} can utilize either CPUs or Processing-in-Memory (PIM) architectures to quickly calculate aggregates to support the visualizations. On real world datasets, we show that \vispim{} generates visualizations that highlight patterns and correlations, some expected and some unexpected.
By using PIM, we can calculate aggregates 45\%-64\% faster than modern CPUs on large datasets. For use cases with 100 million rows and 32 columns, our system is able to compute 4,960 aggregates (each of size 128x128x128) in about a minute.
\end{abstract}


\maketitle

\section{Introduction}
\label{sect-intro}

Data visualization is the most common way of doing data exploration and data science, and is widely used by data analysts to derive useful insights and trends from data. This is shown by the increasing user base of commercial data visual analysis systems such as Tableau~\cite{tableau}. In particular, aggregation queries are a popular candidate for visualizations~\cite{lensxplain, rdfagg, falcon}, as they can help analysts identify correlations and patterns between attributes in the data. As datasets grow increasingly large and complex, there has been a growing interest in interactive visualization of big data, and as a result also in technologies that are scalable and increase the size and complexity of the data our systems can handle~\cite{battlestructreview}.

In this paper, we focus on how to efficiently calculate aggregates on large datasets for the purpose of visualizing two and three dimensional interactions. A two-dimensional visualization will correspond to a
certain kind of heatmap (described in detail below). A three-dimensional visualization is a set of such
heatmaps, corresponding to partitions of the data according to values of the third dimension.
We aim to help users visualize and identify both expected and unexpected
patterns in the data that may be more complex than simple
correlations.

Before describing our approach, we identify desirable features for any such system.
\begin{description}
\item[Scale] The system should scale to large datasets with many rows and many candidate columns.
\item[Focus] Visualizations should focus users' attention to where most of the data resides. One consequence of this
requirement is that pixels should
be allocated roughly in proportion to data density. This requirement is the opposite of what might be required
in systems that aim to identify small numbers of
outliers within a large dataset.
\item[Interaction] Visualizations should focus on the interactions between dimensions, while leaving simpler one-dimensional
patterns to simpler analyses.
\item[Coherence] In a high-dimensional dataset, a system may generate many visualizations for various pairs/triples of
dimensions. The visualization itself should be interpretable in a uniform manner even without looking at the identity of
the dimensions themselves.
\item[Robustness] Many large datasets have a small number of erroneous data items that can confound data visualizations.
For example, an outlier value (perhaps due to an accidental keystroke leading to an extra digit) could skew an axis range so
that much of the visualization space is wasted. Avoiding such problems could require manual data cleaning on many columns.
A visualization system that is robust to a small number of erroneous data items can proceed without this time-consuming
data cleaning step.
\end{description}

We will use the term ``dimension'' to describe an axis of a visualization. 
A dimension may be a column value,
a function applied to some column values, or some lexicographic combination of such values. Each dimension has an implicit order.
For numeric attributes, the numeric ordering is the most natural. For categorical attributes, one can potentially choose multiple
different orderings to get different candidate dimensions. For example, the taxi dataset we use for experiments has a zone-ID string to
identify the pick-up and drop-off zones for a taxi ride. The zone-ID could be ordered (a) alphabetically, (b) by the latitude of
the zone's centroid, (c) by the longitude of the zone's centroid, (d) the lexicographic combination of the borough (e.g., Manhattan)
with latitude or longitude. Each choice would be a different candidate dimension, and induce different clustering patterns on
the corresponding visualizations.

Our visualization scheme is based on {\em independence diagrams}~\cite{BerchtoldJR00,BerchtoldJR98}. 
An independence diagram is a heatmap in which both the horizontal and vertical axes are binned into groups of equal population. By ``normalizing out'' the
one-dimensional distributions, interaction patterns (i.e., adherence to or departures from ``independence'') can be seen
in a manner that devotes visualization space in proportion to data density. When bin counts are large, outliers play a minor role in the definition of axis bins. These properties make the independence diagram a good candidate for the requirements of focus,
interaction and robustness. Coherence is also supported, in the sense that any diagram can be interpreted
in terms of quantiles, e.g., ``there is a cluster in which the top few percent of dimension $x$ are associated with the
bottom few percent of dimension $y$''. Knowing the dimension labels and bin boundaries can make the cluster more concrete, but
the more abstract interaction is visible without that information. There are some technical differences from~\cite{BerchtoldJR00,BerchtoldJR98} in how we determine
bin boundaries, and how we color images; these differences enhance coherence and are described in Sections~\ref{sect-binning}
and~\ref{sect-viz-framework}.

The original independence diagram evaluation~\cite{BerchtoldJR00,BerchtoldJR98} considered only two-dimensional visualizations.
There are ${N \choose 2}$ possible visualizations, each of size $B^2$ where $N$ is the number
of dimensions and $B$ is the number of bins per dimension. For
$N=13$ and $B=100$, the maximum size considered by~\cite{BerchtoldJR00,BerchtoldJR98}, the memory footprint $F$
was sufficiently small that
a single pass through the source data could accumulate aggregates to support all ${13 \choose 2}=78$ diagrams.
To support three dimensional analyses, one needs ${N \choose 3}$ aggregates each of size $B^3$.
For $N=13$ and $B=100$, this footprint is three orders of magnitude larger than the two-dimensional footprint $F$.
For $N=32$ and $B=128$ (a case we'll consider in this paper), the footprint is six orders of magnitude bigger than $F$.
A larger footprint means that
aggregates need to be stored much lower in the memory hierarchy leading to a potential performance bottleneck.
Thus, to address our requirement of scale, we need to devise techniques that can efficiently generate these large aggregate arrays for both general CPUs and Processing-in-Memory devices.
For general CPUs, we implement a method of partitioning updates to aggregates that trades off slow random accesses to memory lower in the memory hierarchy for more sequential scans over the input data. For processing-in-memory devices, we introduce a method that evenly divides aggregates among the many weaker workers in the system without any overlap that is flexible to the number of dimensions in the dataset and the workers in the system, while ensuring that a given tuple in the dataset is only sent to a fraction of the workers, and introduce a fast way of updating aggregates within a worker that minimizes accesses to the main memory while avoiding synchronization primitives.
These techniques constitute the primary contribution of this paper.

Generating so many visualizations creates a new scaling problem, because a user should not be expected to look at
thousands of images, most of which are uninteresting. To handle this problem we allow users to select a heuristic to order
the results, so that the most promising images are presented first. 
\eat{In future work, we plan to analyze the images with
machine learning techniques to automatically identify features that might make an image interesting to a user.}

\eat{
``End-to-end system''~\vispim{} ... \reminder{Is our system a truly end-to-end system?  Could a user download our CPU code and generate visualizations with minimal effort?  Also, we'll need to upload some software artifacts that are suitably anonymized.}
--> Yes, the user can do the above with just the code.
}

Our key contributions are summarized as follows:

\begin{itemize}
    \item We propose DIVAN ("Dimensions Interact - Visualize All Now!"), an end-to-end system that receives as input a dataset, and outputs visualizations corresponding to aggregation queries with 3 group-by columns. \vispim{} uses a binning algorithm that allocates approximately the same number of tuples per bin to automatically focus on high frequency data, while still having fast execution times. 
    \item We implement two versions of~\vispim{}, one that can run on general-purpose CPUs, and one that can run on Processing-in-Memory architectures for better performance. We develop algorithms that utilizes each architecture's characteristics to achieve fast performance.
    \item We show through experiments that~\vispim{} can be effectively used to highlight trends and patterns in real world datasets, and that using a Processing-In-Memory architecture with~\vispim{} can speed up query processing for visualization compared to modern CPU systems.
\end{itemize}

\section{Workflow}

We now describe the workflow that a data analyst might employ to create and analyze visualizations. The first step is to identify a dataset/table of interest. A global set of dimensions would then be specified, based on the columns in the dataset. Some columns may be used in multiple dimensions, using different orders to support different potential clustering patterns. A preprocessing step is then done once on the entire dataset that transforms the original dimensions into integers to speed up later processing steps. This preprocessing is only done once and does not need to be done on subsequent workflows on the same dataset.

At this point the analyst is ready to generate some visualizations. They select a subset of rows to analyze, $N$ candidate dimensions, some number of aggregation functions, and instructs the
system to generate visualizations. The system will compute the aggregates needed for the visualization, transform the results into
images, and compute priorities for the images based on user-selected heuristic rules.  The images will then
be presented to the user in order of priority.  Images may be presented in groups.  For example if dimensions $x$ and $y$ are being
analyzed along with a time dimension $t$, the system may present four $x/y$ images side by side, each corresponding to a
quarter of the dataset partitioned evenly by time.
To increase the diversity of images, we lower the priority of an image (or image group)
if its visualized dimensions $x$ and $y$ have been seen before, with a different partitioning dimension.

Based on the results, the user may choose to analyze the dataset with a new combination of dimensions
and aggregate functions.
The user always has the option of returning to the original database to get finer-granularity results
for specific visualizations.

\section{Dataset Normalization}

In this section, we describe how we bin dimensions of the dataset of interest in a normalized fashion so that each bin contains an equal number of tuples.

\subsection{Binning via sorting}
\label{sect-binning}

For each dimension in the dataset, a simple way to achieve the aforementioned binning is to first sort the dataset according to the dimension of interest, and then assign bins based on the sort order. For example, if 100 bins are used, the first 1\% of tuples in the sorted dataset will be assigned the first bin, and the last 1\% of tuples in the sorted dataset will be assigned the hundredth bin.

Binning the dataset this way allocates bins to values roughly proportional to the frequency of the value. For example, a value that appears in 30\% of the dataset will be allocated around 30\% of the total bins. The same principle applies to continuous domains as well, such as time. If 30\% of the dataset falls in the range of 12pm to 1pm, 30\% of the total bins will be allocated to values within that range. This translates to more pixels being allocated to more frequent values/ranges in the resulting visualization, which helps the user focus on where most of the data resides. This also makes \vispim{} robust, as it prevents outliers from skewing an axis range on a visualization by allocating fewer bins to infrequent values.

Note that different values can be mapped to the same bin, such as when infrequent values that do not have enough tuples to fill a bin share their allocated bin with other values, and when a value fills the leftover space in the previous bin. This approach preserves the even spacing of bins in pixel space, at the cost of possibly ``diluting'' visual patterns as correlations associated with different values are grouped into the same visualization space, i.e. pixels. However, this is a minor issue as frequent values are mapped to many bins and are thus less affected by sharing up to two bins with other values, and infrequent values that are greatly affected by this dilution are mapped to few bins and are not the focus of the visualization. A higher linear image resolution, proportional to the number of bins, can also reduce the effects
of dilution.

While our binning method is related to equidepth binning, it differs from it due to the fact that a single value may be mapped to multiple bins, which allows \vispim{} to dedicate more visualization space to frequent values. This approach differs from independence diagrams~\cite{BerchtoldJR00,BerchtoldJR98} which don't split single values across multiple bins. Instead,
independence diagrams scale the visualization pixel width of a bin based on bin frequency.

\subsection{Approximate binning via histograms}

The aforementioned method of binning is a performance bottleneck in the workflow as it is required to sort the entire dataset for each dimension every time one wishes to perform data analysis, which hurts \vispim{}'s scalability.

For the purposes of visualization, we observed that binning does not have to be perfect, and having small variances in the number of tuples in each bin has little effect on visualizing the patterns present in the data. In this section, we describe how we can introduce a preprocessing step to speed up the binning process on all subsequent analysis workflows on the dataset, regardless of the subset of the dataset used for analysis.

\subsubsection{Dataset preprocessing}

We preprocess the dataset by doing an argsort for every dimension in our dataset, which results in an additional integer column that corresponds to the sorted index of the tuple. This preprocesssing step is only done once, and the resulting augmented dataset can be used for all subsequent analysis workflows on any subset of the dataset. By doing preprocessing, we speed up subsequent analysis workflows as the dimension domains are converted into integers, and also supports scenarios where the analyst may generate additional visualizations focusing on areas of interest, and also scenarios where multiple analysts may perform data analysis on different subsets of the same dataset.

\subsubsection{Dataset binning}

At the time of visualization, for each dimension the dataset is approximately binned by creating a histogram with a number of bins much larger than the target number of bins used for analysis. For example,
we might create a histogram with 1048576 bins with the purpose of subsequently re-binning the histogram into 128 bins. The following pseudocode describes the process of binning the dataset for a single dimension, using a histogram with 1048576 bins.

\begin{Verbatim}[frame=single, fontsize=\fontsize{7.4pt}{10pt}\selectfont]
Input: data[],            // Sorted indexes of dimension
       num_tuples,        // Number of tuples in dataset subset
       total_num_tuples,  // Number of tuples in dataset
       bins               // Number of bins used for binning
Output: binned_data[]     // Array to store binned dataset

// Calculate number of bits needed to 
// represent a sorted index in the data set.
idx_bits = ceil(log_2(total_num_tuples))

// Allocate zeroed out space for histogram
HISTO_SIZE = 1024576 // 2^20
int histogram[HISTO_SIZE]

// Use the upper 20 bits of each sorted idx 
// to index into and update the histogram.
shift = idx_bits - 20
for (i = 0; i < num_tuples; i+=1):
    histogram[data[i] >> shift] += 1

// Iterate through histogram and replace bin contents with
// bin index based on the number of tuples seen before.
tuples_per_bin = ceil(num_tuples / bins)
seen = 0
for (i = 0; i < HISTO_SIZE; i+=1):
    num_in_bucket = histogram[i]
    histogram[i] = seen / tuples_per_bin
    seen += num_in_bucket

// Allocate space to store binned dataset
int binned_data[num_tuples]

// Use the upper 20 bits of each sorted idx to index
// into the histogram and get the corresponding bin index.
for (i = 0; i < num_tuples; i++):
    binned_data[i] = histogram[data[i] >> shift]
return binned_data
\end{Verbatim}

Note that this approach is not effective for datasets that are very small subsets of the preprocessed dataset, in which case we can get heavily skewed histograms that skew the number of tuples in each resulting bin. Using 128 bins for binning and using histograms with $2^{20}$ bins, we encounter skew when using subsets of the dataset are smaller than $1 / 2^{13}$th of the preprocessed dataset. However, this is not a big issue as even for large datasets with a 100M rows, $1 / 2^{13}$ is only 12k rows, at which point one can quickly sort the rows directly.

Our binning scheme can also handle small inserts. As it is inefficient to preprocess the data every time a batch of new rows are inserted, we can handle small inserts by approximately keeping track of the histogram's bucket boundaries by saving the original domain value of the first tuple to enter each bucket. Then, we can iterate over the new tuples and update the histogram using the saved boundaries, and then assign bins to the new rows using the histogram. While this code path is slow compared to the code path for preprocessed indexes, the time to process the new rows is only a fraction of the total binning time if the batch of new rows is small compared to the dataset of interest.


As frequent values in the dataset are assigned more bins, there is potential to extract more insights from the dataset if we assign bins for frequent values based on secondary attributes, as we can use the bins allocated to a frequent value to further divide the data within a range of visualization space. We show examples of this in Section \ref{s:vis_comp}.

\section{Aggregate calculation}

After binning the data using $B$ bins, we now have a dataset consisting of $N$ columns of binned data, derived from the dimensions of the original dataset, along with an additional column corresponding to the value to be aggregated. All of the $N$ binned columns have an approximately even distribution of $B$ distinct values. For our visualization system, our goal is to calculate all GROUP-BY queries with every combination of three GROUP-BY columns, which is a total of ${N \choose 3} * B^3$ aggregates.

\eat{
\reminder{I think we can omit the following paragraph.}
Each tuple in the binned dataset is used to update ${N \choose 3}$ locations in memory. While such calculation of aggregates can be easily implemented for a general-purpose CPU, due to the high volume of data being written to many memory locations we found that performance is bottlenecked by the CPU's memory bandwidth. To achieve better performance and thus better interactive times, we found that an implementation that utilizes Processing-In-Memory (PIM) \reminder{(TODO: cite PIM here)}, an emerging technique that aims to reduce costly data movement, can help accelerate computation of the aggregates. In the following sections we explain how we implement the computation of aggregates on both a general-purpose CPU and a commercial PIM system prototype from UPMEM \reminder{(TODO: cite UPMEM)}.
}

\subsection{General CPU Execution}

We assume that the input is stored in columnar fashion,
and that the aggregate CPU memory is large enough to hold the entire output.
A straightforward way to compute the aggregations is as follows:
\begin{verbatim}
for r = 1 to #records
  read record r
  for j = 1 to #aggregations
    update agg #j using r
\end{verbatim}
One problem with the method above is that it has poor locality. Every aggregation update
is likely to be a miss in the lowest level CPU cache.  A second problem is that if we attempt
to distribute work to multiple threads by parallelizing the outer loop (partitioning the input)
then costly synchronization primitives are necessary to prevent interference.

An alternative way to compute the aggregations changes the loop order:
\begin{verbatim}
for j = 1 to #aggregations
  for r = 1 to #records
    read record r
    update agg #j using r
\end{verbatim}
At first glance this rewritten version seems inefficient because the input needs to be scanned
multiple times.  However, the input reading cost is manageable because (a) one only needs to read the columns
involved in the current aggregation, and (b) hardware prefetching hides the memory latency associated with
sequential access of the input. This method has better temporal locality because there is repeated access to a single
aggregate array rather than interleaved access to all aggregate arrays. Further, parallelization of the outer loop can proceed without synchronization
because the output for each thread is disjoint. 
We can further explore the tradeoff between temporal locality and scans through the input by dividing a single aggregate array into smaller partitions, and doing a scan for each partition.
As we shall see experimentally in Section~\ref{sect-exp-CPU}, dividing the aggregates into smaller partitions may be
beneficial when an individual aggregation alone has a large memory footprint.

\eat{
Multithreaded execution using CPUs

Two methods: Simply write directly to memory (large memory footprint) or partition aggregation memory and write to partitions that fit in memory (more passes over data, but better cache locality)
}

\vspace{-0.5mm}

\subsection{Processing-In-Memory Architecture}

\eat{
Each tuple in a dataset with $N$ dimensions needs to update $N \choose 3$ locations in memory. This becomes a memory bottleneck if there is a large number aggregates. In this section, we detail how a processing-in-memory system can be used to mitigate this memory bottleneck and calculate aggregates more quickly.
}

Processing-in-memory (PIM) refers to the concept of being able to execute operations inside or near the memory of a system, which reduces costly data movement to and from the main CPU. While the concept has been around for some time~\cite{MUTLU201928, mutlu2022modernprimerprocessingmemory, gomeznewpara}, only recently has advances in technology made PIM available on commerical hardware. In particular, the UPMEM PIM architecture~\cite{upmemthefirst} is the first PIM system that is commercially available on real hardware, and is the system we have used in this paper.

 The UPMEM PIM system consists of a single monolithic CPU, which we will refer to as the \textit{host CPU}, and many weaker processing units integrated with the system's DRAM arrays, called DRAM processing units (DPUs). The host CPU can send and receive data from the DPUs, and can signal to the DPUs to run their code. It is possible for the host CPU to signal the DPUs in either synchronous mode, where the host CPU is blocked until all the DPUs finish, or asynchronous mode, where control is immediately returned to the host CPU right after signalling the DPUs. Direct communication between DPUs is not possible in this model, and all data sent and received by the DPUs must be through the host CPU.

 A DPU is a weak, general purpose processor running at 450 MHz. It is capable of running C code, and consists of a single core that has 24 threads, 64MB of MRAM which serves as its main memory and 64KB of WRAM that serves as its cache~\cite{upmemtech}. Despite its low processing power, a PIM system has the potential to run data intensive workloads quickly by utilizing the thousands of DPUs equipped on the system in parallel. Because of this it is of utmost importance to achieve workload balance for a task involving PIM, and minimize data movement between the host CPU and DPUs.

\subsection{Aggregation Distribution}\label{s:agg_dist}

Given a preprocessed dataset with $N$ dimensions, \vispim{} needs to calculate $N \choose 3$ aggregates, where each aggregate is a group by between 3 different dimensions in the dataset. To efficiently do this with a PIM system, these aggregates need to be distributed evenly across all DPUs so that each DPU does an equal amount of work.

A simple way to distribute aggregates is to send all aggregates to all DPUs. We can then distribute the tuples (each of which is an entire row of $N+1$ columns) in the dataset evenly across all DPUs for the DPUs to read and use to update the aggregates. However, since each DPU only has 64MB of MRAM to hold aggregates, this method is infeasible even for datasets that only have 8 dimensions.

Another way of distributing aggregates is to evenly distribute the $N \choose 3$ aggregates among DPUs. Although this does not raise storage issues as each DPU now holds a disjoint subset of the aggregates, since each tuple is used to update every aggregate, every tuple needs to be sent to every DPU, which results in a performance bottleneck due to data transfer between the host CPU and DPUs.

Instead, we can do the following. Suppose our dataset has $N$ dimensions, each of which is binned using $B$ bins. We first divide the $N \choose 3$ aggregates into $N$ equal sized groups, where all aggregates in group $i$ share dimension $i$.\footnote{This is always possible when $N$ is not a multiple of 3; we handle multiples of
3 slightly differently.}  We further divide group $i$ into $B$ different subgroups by partitioning all the aggregates in group $i$ by distinct bins in dimension $i$. Each subgroup is then sent to a different DPU. With this aggregation distribution, each DPU holds a disjoint subset of the aggregates, while we get reasonable transfer times as each tuple only needs to be sent to $N$ distinct DPUs since a tuple only updates one subgroup for every group. Furthermore, our approximate binning strategy ensures that we achieve almost perfect workload balance across DPUs as roughly equal numbers of tuples are sent to each subgroup. However, with this scheme, we utilize exactly $N * B$ DPUs, which may be different from the actual number of DPUs in the system. We now
explain how to extend our scheme to an arbitrary number of DPUs to the nearest multiple of $B$, and go into detail on how we exactly distribute aggregates.

We index the dimensions using integers $0...N-1$. We represent an aggregate between three dimensions $d_0, d_1, d_2$ as $(d_0, d_1, d_2)$, where $d_0, d_1, d_2$ are distinct integers in the range $0...N-1$.

\begin{definition} [Shift]
For an aggregate $(d_0, d_1, d_2)$, and an integer $s$, we define $\shift((d_0, d_1, d_2), s)$ as $((d_0 + s) \% N, (d_1 + s) \% N, (d_2 + s) \% N)$.
\end{definition}

\begin{definition} [Equality]
We say two aggregates $(d_0, d_1, d_2)$, and $(e_0, e_1, e_2)$ are equal ($(d_0, d_1, d_2) = (e_0, e_1, e_2)$) if the set of dimensions for both aggregates are equal (i.e. $\{d_0, d_1, d_2\} = \{e_0, e_1, e_2\}$)
\end{definition}

Consider first the case where $N$ is not a multiple of 3.
To distribute $N \choose 3$ aggregates into $N$ equal-sized groups, where all aggregates in group $i$ share dimension $i$, we first find ${N \choose 3} / N$ aggregates that belong in group 0. Next, to get aggregates in group $i$, where $1 \leq i < N$, for every aggregate $A$ in group 0 we apply $\shift(A, i)$. This requires that the initial aggregates in group 0 are chosen carefully, so that we do not get duplicate aggregates (and thus miss other aggregates) when calculating aggregates for other groups.

Note that we get duplicate aggregates when creating aggregates for other groups when there exists aggregates in group 0 that are equal with each other after a shift operation. For example, if $N = 5$ and group 0 contains $(0, 1, 2)$ and $(0, 1, 4)$, since $\shift((0, 1, 2), 4) = (4, 0, 1) = (0, 1, 4)$, we get duplicate aggregates when creating aggregates for group 4. To find aggregates for group 0, our goal then, is to find ${N \choose 3} / N$ aggregates which satisfy the conditions that all aggregates contain dimension 0, and that for any two different aggregates $A_0$ and $A_1$, $\shift(A_0, s)$ is not equal to $A_1$ for any $s$.

\begin{definition} [Shift overlap]
We say two aggregates $A_1$ and $A_2$ \textit{shift overlap} if there exists $s$ where $\shift(A_1, s) = A_2$.
\end{definition}

Without loss of generality, an aggregate in group 0 can be represented as $(0, a, b)$, where $0 < a < b < N$. 
\eat{
Except in the special case where $N \% 3 = 0$ and $a = N / 3$ and $b = 2 * N / 3$ (which we show how to handle later),}
A given aggregate $A = (0, a, b)$ shift overlaps with two other aggregates that contain dimension 0, namely $\shift(A, N - a)$ and $\shift(A, N - b)$. Because there are ${N \choose 2} = (N-2)(N-1)/2$ possible combinations of $a$ and $b$ where $0 < a < b < N$, we need to choose the third of these aggregates, which in total is $(N-2)(N-1)/6 = {N \choose 3} / N$ aggregates, which do not shift overlap with each other. We can do this by iterating through each possible $a$ and $b$, to get aggregate $A = (0, a, b)$, perform $\shift(A, N - a)$ and $\shift(A, N - b)$ and sort them respectively, and only choose to put $A$ in group 0 if $A$ is the first in the lexicographical order between the 3 aggregates generated. Algorithm~\ref{a:group0} describes how to obtain group 0.

\begin{algorithm}
\caption{Create group 0}\label{alg:cap}
\begin{algorithmic}[1]
\Procedure{Group0}{$N$} \Comment{N: Number of dimensions}
\State group0 = []
\For{a in 1,2,..N-2}
\For{b in a+1,a+2..N-1}
\If{N \% 3 = 0 \textbf{and} a = N / 3 \textbf{and} b = (N / 3) * 2}
\State\textbf{continue}
\EndIf
\State A = (0, a, b)
\State overlap1 = sort(shift(A, N-a))
\State overlap2 = sort(shift(A, N-b))

\If{A < overlap1 \textbf{and} A < overlap2}  
    \State group0.add(A)
\EndIf
\EndFor
\EndFor
\State \textbf{return} group0
\EndProcedure
\end{algorithmic}
\label{a:group0}
\end{algorithm}

Once we create group 0, we can create groups 1 to N-1 by shifting aggregates in group 0. This process is described in Algorithm~\ref{a:evendist}. When $N$ is a multiple of 3, ${N \choose 3} / N$ is not an integer,
and we cannot shift aggregate $(0, N / 3, (N / 3) * 2)$ to group $i$ where $i \geq N /3$ without duplicating aggregates. In this case, for only group i where $i < N / 3$ we add $\shift((0, N / 3, (N / 3) * 2), i)$ to the group. As a result when $N$ is a multiple of 3 we still have $N$ groups, but some of the groups compute an extra aggregate relative to others.\footnote{The performance impact of this imbalance is minor in practice; for $N=24$ some groups will have 84 aggregates and some will have 85.}

\begin{algorithm}
\caption{Distribute $N \choose 3$ 3D aggregates into N groups where each group contains a common dimension}\label{alg:cap}
\begin{algorithmic}[1]
\Procedure{EvenDist3D}{$N$}\Comment{N: Number of dimensions}
\State group0 = Group0(N)
\State groups = [group0]
\For{i in 1,2,..N-1}
    \State group\_i = []
    \For{agg in group0}
        \State group\_i.add(shift(agg, i))
    \EndFor
    \If{N \% 3 = 0 \textbf{and} i < N / 3}
        \State group\_i.add(shift((0, N/3, 2*(N/3), i))
        \EndIf
    \State groups.add(group\_i)
\EndFor
\State \textbf{return} groups
\EndProcedure
\end{algorithmic}
\label{a:evendist}
\end{algorithm}

Using the aforementioned procedures, it is possible to effectively utilize $N * B$ DPUs by creating $N$ groups and partitioning the groups into $B$ subgroups, as explained above. However, suppose the system has fewer than $N * B$ DPUs. Because a group is divided among $B$ DPUs, let's say our system has $R * B$ DPUs, where $R < N$. In this case, it is possible to calculate the aggregates separately by dividing the $N \choose 3$ aggregates into one group that contains aggregates that contain at least one dimension from $0..R-1$ and one group that contains aggregates that only contains dimensions from $R..N-1$, and treat them as separate problems that require different aggregation distributions. For the first group, we can create $R$ groups where each group contains a common dimension by using Algorithm~\ref{a:evendist} to distribute aggregates that contain at least one
dimension from $0..R-1$, and distribute aggregates that contain only dimensions from $R..{N-1}$ to groups $0..R-1$ by utilizing another function, \textit{EvenDist2D}, that distributes $N \choose 2$ 2D indexes into N groups where each group contains a common index. This process is explained in Algorithm~\ref{a:evendist2d} and Algorithm~\ref{a:split}. For the second group, since we are dealing with ${N - R} \choose 3$ aggregates that only contain dimensions from $R..N-1$, we can recursively handle this as a problem of calculating aggregates for a dataset with $N-R$ dimensions. This recursive process is explained in Algorithm~\ref{a:even_dist_recursive}, and \vispim{} calculates groups for each iteration of the recursion by calling $DpuDist(N, B, D, 0)$, where $D$ is the number of DPUs in the system. Note that in the case where the number of DPUs is not a multiple of $B$, \vispim{} utilizes the nearest multiple of $B$ DPUs.

\begin{algorithm}
\caption{Distribute $N \choose 2$ 2D indexes into N groups where each group contains a common index}\label{alg:cap}
\begin{algorithmic}[1]
\State \(\triangleright\) N: Number of indexes
\State \(\triangleright\) front: Determines whether front half of groups or back half of groups get more aggregates when N \% 2 = 0
\Procedure{EvenDist2D}{$N$, front}

\State groups = []
\For{i in 0,1,..N}
\State num\_aggs = $\floor{{N \choose 2} / N}$
\If {($i < N / 2$ \textbf{and} front) \textbf{or} ($i \geq N / 2$ \textbf{and} !front)}
\State num\_aggs += 1
\EndIf
\State group\_i = []
\For {j in 1,2,..num\_aggs}
\State group\_i.add((i, (i + j) \% N))
\EndFor
\State groups.add(group\_i)
\EndFor
\State \textbf{return} groups
\EndProcedure
\end{algorithmic}
\label{a:evendist2d}
\end{algorithm}

\begin{algorithm}
\caption{From $N \choose 3$ 3D aggregates, distribute aggregates that contain at least one dimension from 0..R-1 into R groups where each group contains a common dimension}\label{alg:cap}
\begin{algorithmic}[1]
\State \(\triangleright\) N: Number of dimensions, R: Number of groups
\Procedure{Split}{$N, R$}

\If {$R \geq N$}
\State \textbf{return} EvenDist3D(N)
\EndIf

\State groups = EvenDist3D(R)

\State \(\triangleright\) Distribute aggregates that contain two dimensions in R,R+1,..N-1

\For{dim in 0,1,..R-1}
    \For{i in R,R+1,..N-2}
        \For{j in i+1,i+2,..N-1}
            \State groups[dim].add((dim, i, j))
        \EndFor
    \EndFor
\EndFor

\State \(\triangleright\) Need to evenly use both 2d groups to get even distribution
\State 2d\_groups\_front = EvenDist2D(R, True)
\State 2d\_groups\_back = EvenDist2D(R, False)

\State \(\triangleright\) Distribute aggregates that contains one dimension in R,R+1,..N-1
\State curr = 0
\For{dim in R,R+1,..N-1}
    \If{curr \% 2 = 0}
        \State 2d\_groups = 2d\_groups\_back
    \Else
        \State 2d\_groups = 2d\_groups\_front
    \EndIf
    \For{i in 0,1,..R-1}
        \For{g in 2d\_groups[i]}
            \State groups[i].add((g[0], g[1], dim))
        \EndFor
    \EndFor
    \State curr += 1
\EndFor

\State \textbf{return} groups
\EndProcedure
\end{algorithmic}
\label{a:split}
\end{algorithm}

\begin{algorithm}
\caption{Distribute $N \choose 3$ aggregates binned with B bins to D DPUs, using multiple iterations}\label{alg:cap}
\begin{algorithmic}[1]
\State \(\triangleright\) N: Number of dimensions, B: bins,
\State \(\triangleright\) B: bins, S: Offset to shift aggregates
\Procedure{DpuDist}{$N, B, D, S$}
\State R = int($D / B$) \Comment{Number of DPU `rows' of B DPUs}

\State it = Split(N, R)
\For{group in it}
    \For{agg in group}
        \State agg = shift(agg, S)
    \EndFor
\EndFor
\State iterations = [it]
\If {N - R > 2}
    \State iterations += DpuDist(N-R, B, D, R)
\EndIf

\State \textbf{return} iterations

\EndProcedure
\end{algorithmic}
\label{a:even_dist_recursive}
\end{algorithm}

\eat{
\reminder{If we have space, perhaps put in a small example?}
}

\subsection{Execution}

The following sections detail the roles of the host CPU and DPUs.

\subsubsection{Host CPU}\label{s:host cpu execution}

Before the actual aggregate calculation, \vispim{} uses Algorithm~\ref{a:even_dist_recursive} to evenly distribute aggregates to the DPUs. Each aggregate group in each iteration returned by Algorithm~\ref{a:even_dist_recursive} is divided into $B$ different subgroups by partitioning the aggregates in each group by distinct bins in the common dimension of the group. The dimension indexes in each subgroup is then sent to each DPU, in the form of a contiguous integer array.

Which subgroup a DPU is in charge of updating determines which tuples are sent to it. For example, if a subgroup is the 6th subgroup in a group where dimension 5 is the common dimension, the corresponding DPU receives tuples whose 5th dimension has the value 6. For the purpose of sending tuples to the DPUs for aggregation calculation, the host CPU allocates a buffer of 320KB for each DPU. The host CPU then iterates through the data and fills the buffers with the tuples the DPUs need. If an iteration contains R groups, a tuple is sent to R different buffers, one for each group. During this step, the host CPU utilizes multithreading and allocates one thread to each group to fill the buffers faster.

For an iteration, in the case we have two or more times more DPUs than needed to calculate aggregates, we duplicate the aggregates assigned to each DPU to the leftover DPUs, and partition tuples evenly among the DPUs that share the same aggregates so that each DPU does less work. Since the aggregates are duplicated, the host CPU needs to do a final aggregation of the duplicated aggregates at the end of execution. However, as this final aggregation only depends on the number of dimensions and number of bins, for large datasets this step has little effect on performance.

Once a single buffer in a group is filled with tuples, the corresponding thread stops execution and waits for other threads to stop. Due to the properties of our binning strategy, all buffers fill at approximately the same rate, which leads to good load balance. Once all threads stop execution, the main thread initiates a DPU transfer, and sends the contents of all buffers to their corresponding DPUs. Once the transfer is over, the host CPU signals the DPUs to start running their code and calculate aggregates. Once aggregate calculation is done, the host CPU repeats the above process for the remaining tuples in the data. In asynchronous mode, control is returned to the host CPU as soon as the host CPU signals the DPU to run their code, which allows \vispim{} to overlap host CPU buffer filling and DPU execution.

At the end of execution, the host CPU initiates a data transfer from the DPUs to the CPUs, to read back the calculated aggregates to be used for visualization.

\subsubsection{DPU processing}

For every tuple, a DPU needs to update a different location in memory for every aggregate in its subgroup. A simple way to update aggregates is to evenly distribute the $A$ aggregates across the threads in the DPU, and update each aggregate individually. However, we found that this was slow as each thread needed to do an expensive scan over the tuples in MRAM.
Instead, we introduced an optimization that takes advantage of the DPU's ability to directly manage WRAM. Aggregation updates are now done in two phases. In the first phase, all threads work together and read 32KB worth of tuples from the MRAM into WRAM. In the second phase, the threads work independently and iterate through the tuples in WRAM and update their assigned aggregates. This improved execution time because as a whole, the DPU only needs to do one pass through the tuples stored in MRAM, as opposed to doing a pass in MRAM for every thread.

\section{Transforming Aggregates to Visualizations}
\label{sect-viz-framework}

\begin{figure}
    \centering
    \includegraphics[width=0.8\columnwidth]{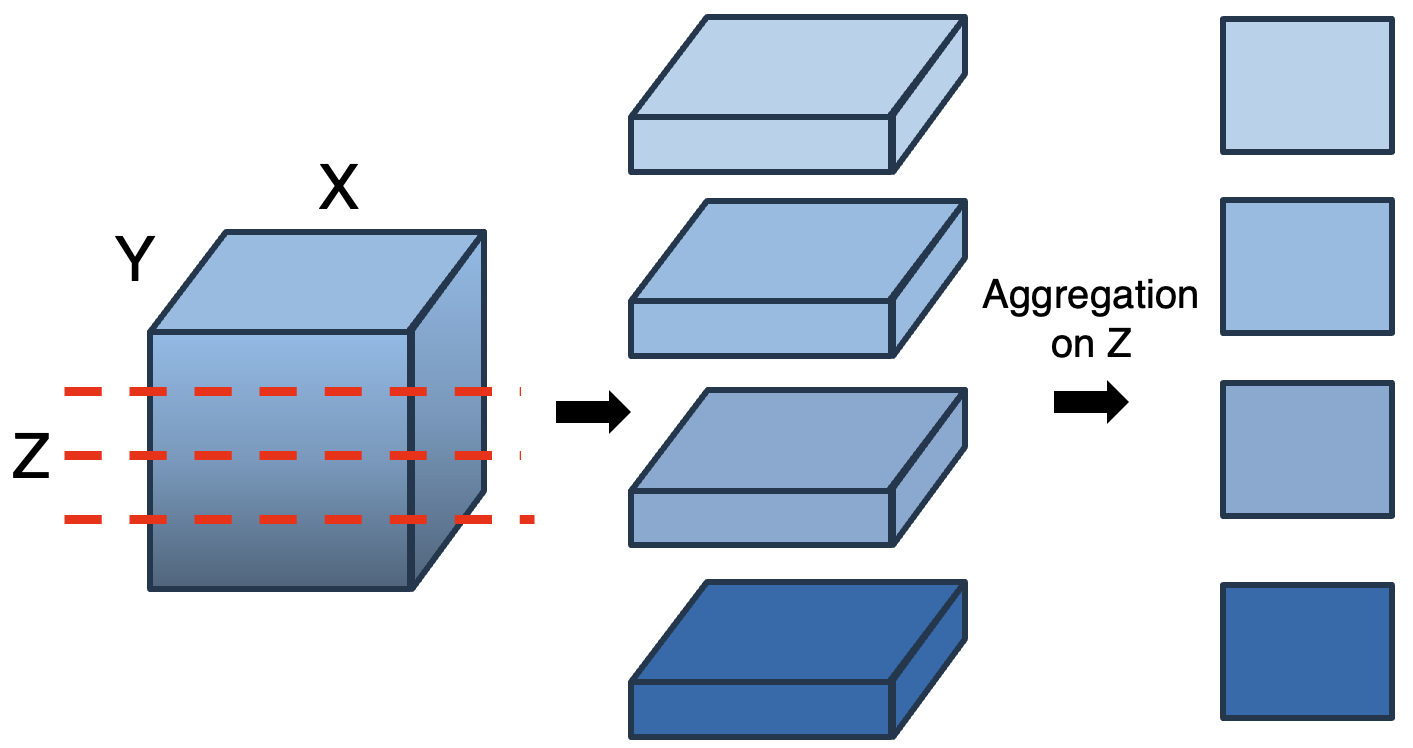}
    \caption{Process of creating 2D images from 3D aggregates}
    \label{fig:agg_process}
\end{figure}

For a given visualization, we use $x$ and $y$ to refer to the horizontal and vertical dimensions of the image and $z$
to refer to the partitioning dimension. A subset of the data can be defined by a contiguous range of $z$ slices.
Data within the region is aggregated to generate totals grouped by $x$ and $y$. A visual representation of this process is shown in Figure~\ref{fig:agg_process}.
Any conventional aggregate functions can be used as the basis for determining pixel intensities via an intensity function
$I$ mapping an aggregate value to (red,green,blue) triples. Pixel intensities vary from 0 to 1 in each
component.

For most of the diagrams in this paper, we define $I$ to be a function that compares the aggregate value with the expected value
in each cell. By construction, when the $x$ and $y$ dimensions are independent, each cell
will have roughly the same aggregate subtotal $S$, equal to the total for the whole dataset divided by $B^2$, the number of visualized
cells. We will color cells red when their totals are larger than $S$ (hot areas that are over-represented),
blue when their totals are smaller than $S$ (cold areas that are under-represented), and black when their totals are
close to $S$ (neutral areas that have the expected contribution). We avoid the use of green because it can be confused with
red by colorblind individuals.
So $I(v,S)$ is given by
\begin{itemize}
    \item $(0,0,1-v/S)$ when $v \leq S$
    \item $(min(1,v/S-1),0,0)$ when $v > S$
\end{itemize}
This design has the feature that brightness increases as the distribution departs from expectation.
The most uninteresting images (or regions within images) are therefore the blackest images/regions and they are easy to ignore.
The positive intensities saturate at twice the expected value. Without saturation, one could alternatively
scale to the maximum intensity of any cell. However scaling in this fashion
would (a) dim the intensities of many other cells in the visualization and be contrary to the {\em focus} requirement, and (b) create a data-dependent intensity scale, contrary to the {\em coherence} requirement; see Section~\ref{sect-intro}.

As originally proposed~\cite{BerchtoldJR00,BerchtoldJR98}, independence diagrams scale the intensity in grayscale between the 5th and 95th percentiles of
the cell value distribution, which can cause problems for {\em focus} and {\em coherence} for some data distributions.
Further, brightness does not necessarily increase as the distribution departs from expectation, making it harder to separate
interesting from uninteresting regions.

\eat{
\reminder{include a discussion of how $z$ is also split if there are ties at the boundary}

\reminder{Show one or two simple examples}
}

There is a trade-off in the choice of $B$, the number of bins in each dimension. Doubling $B$ doubles the linear resolution
of the generated visualizations, while multiplying the space requirement for the output by $2^3=8$. One should therefore seek a
sweet-spot that provides sufficient resolution to see interesting patterns while achieving a feasible memory footprint.
A value of $B\geq 100$ is sufficient to see patterns that affect at least 1\% of the data.
\eat{\textcolor{blue}{, as those patterns will span multiple bins as each bin will contain less than or equal to 1\% of the data.}.}

\section{Visualization Recommendation}\label{s:vis recommendation}

To present interesting images to the user, \vispim{} scores images and presents them to the user in order of score. \vispim{} computes image scores using the average of the red values of all RGB pixels. This heuristic is chosen so that a high score is given to images that contain a lot of red (i.e., where the over-represented data is) across a large area of space. \vispim{} then groups together every image that shares the same $x$ and $y$ axes, and then scores each group by taking the sum of scores of all images belonging to the group. The user is then presented with the top $n$ images of the top $m$ groups, where $n$ and $m$ are configurable. By using a small $n$, the user can enhance the diversity of images they see.

\section{Experimental Evaluation}\label{s:experiment results}

To demonstrate the usage and usefulness of our system, we evaluate \vispim{} on real-world datasets in this section. We focus on two aspects, its performance in calculating aggregates and the visualization images it generates.

\subsection{Dataset Description}

We use two real word datasets in our evaluation. The first dataset, which we will refer to as the taxi dataset~\cite{nyctaxidataset}, is a dataset that details taxi trips in New York during 2019 and 2020. The dataset contains around 100 million rows, where each row corresponds to a single taxi trip. After joining the dataset with a lookup table that contains additional information about taxi zones, the dataset contains 28 columns that describe each trip with characteristics such as pickup/dropoff location, trip distance, passenger count etc, which we each use as a dimension in our analysis. We create more dimensions for analysis by choosing the taxi service zone, latitude and longitude columns as secondary columns to pickup/dropoff locations and passenger count columns, for a total of 32 dimensions to inspect during analysis. To test various aggregate data types, we use counts (32-bit integers), fare amount (32-bit floats) and trip distance (64-bit doubles) as values to aggregate.

Our second dataset, which we will refer to as the flight dataset~\cite{flightdataset}, is a dataset that details global flights from 2018 and 2022. The dataset contains around 29 million rows, where each row corresponds to a single flight. The dataset contains 61 columns that describe the flights, such as flight date, origin/destination, delay time, etc. However, many of the columns have a 1:1 relationship between each other (e.g. origin airport and origin city), so for our experiments we pick out and use 24 columns to use as dimensions that do not have a 1:1 relationship with each other. For aggregate values, we use counts (32-bit integers), arrival delay (casted to 32-bit floats) and departure delay (64-bit doubles).

\subsection{Experiment machines}

For experiments using Processing-in-Memory, we use a server provided by UPMEM. The server is equipped with two Intel Xeon 4215 CPUs, each with 8 cores and 2 threads per core. The server is equipped with 20 UPMEM DIMMs for a total of 2560 DPUs and 160GB memory. However, at the time of experiments only 2210 DPUs were available due to technical issues, so we use 2048 DPUs in our experiments.

For experiments on general CPUs, we use a server equipped with two Intel Platinum 8488C CPUs (the most recent general purpose Intel CPU available on AWS at the time of experiments), each with 48 cores and 2 threads per core. There is 768GB of available memory.

\subsection{Varying Dimensions and Data Types}

\begin{figure*}[!t]
    \centering
    \includegraphics[width=0.9\textwidth]{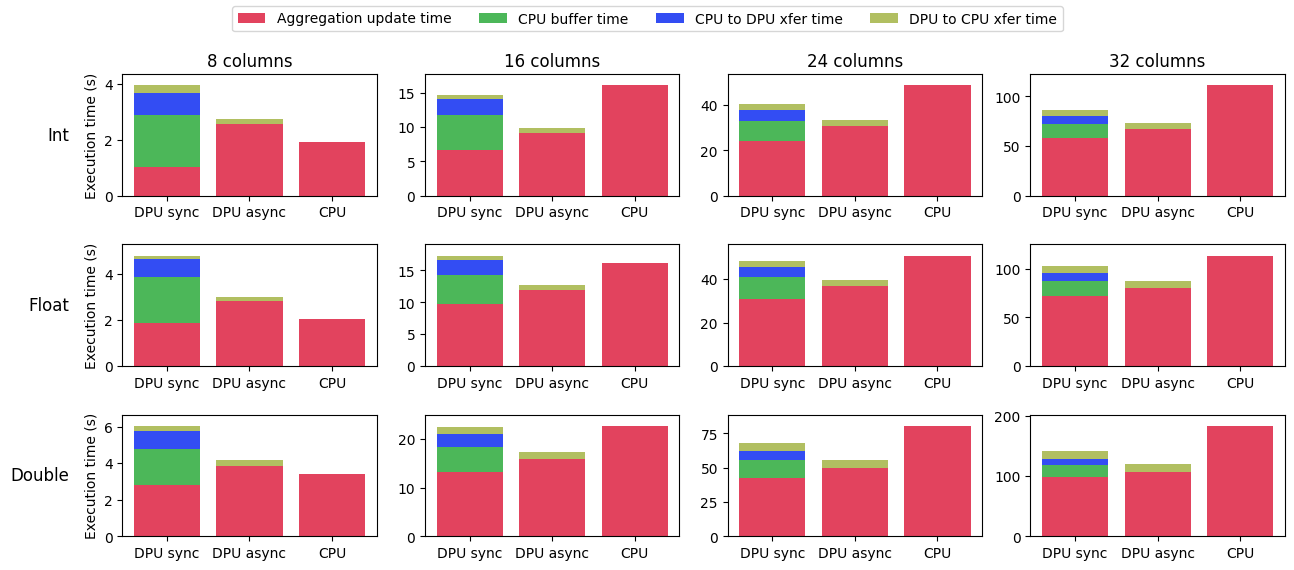}
    \caption{Execution time for varying number of dimensions (Taxi dataset)}
    \label{fig:col_test_taxi}
\end{figure*}

\begin{figure*}[!t]
    \centering
    \includegraphics[width=0.9\textwidth]{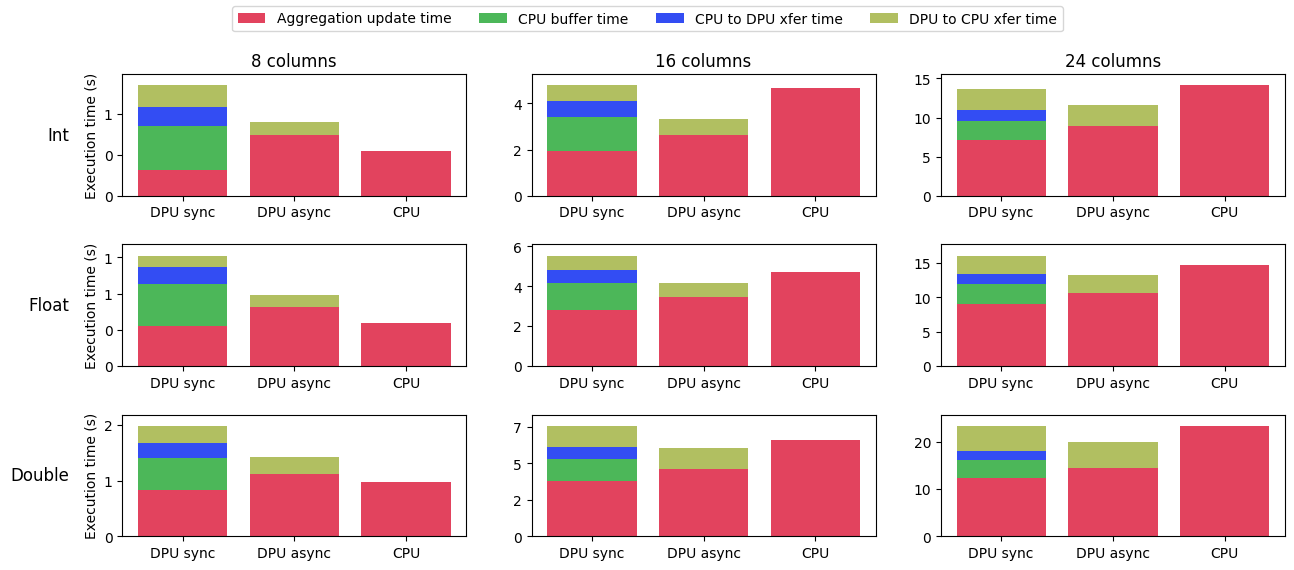}
    \caption{Execution time for varying number of dimensions (Flight dataset)}
    \label{fig:col_test_flight}
\end{figure*}

\begin{figure}
    \centering
    \includegraphics[scale=0.5]{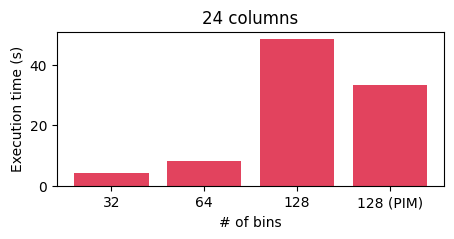}
    \caption{Execution time for various numbers of bins}
    \label{fig:bin_test}
\end{figure}

We evaluate the time taken to compute aggregates on real world datasets, varying the number of dimensions and aggregate value data types. For the CPU machine, we run the aggregation calculation method (Section ~\ref{s:host cpu execution}) using all 192 threads, and run individual experiments where we divide the aggregate arrays into 1, 2, and 4 partitions and report the best execution time. For all experiments, we bin the data using 128 bins. 
For the PIM system, we run experiments for both synchronous mode and asynchronous mode. For synchronous mode, we report the time taken to calculate and write aggregates in the DPUs, fill buffers in the host CPU, and transfer data to and from the host CPU. For asynchronous mode, we report the total time it takes to calculate the aggregates, along with the time to transfer aggregate data from the DPUs to the host CPU.

For the taxi dataset, we measure the time taken to calculate aggregates using 8, 16, 24 and 32 dimensions on aggregate values of type int, float, and double. Results are shown in Figure~\ref{fig:col_test_taxi}. For 8 dimensions, the computation is close to interactive,
taking a few seconds to compute ${8 \choose 3}=56$ aggregates each of size $2^{21}$ elements. For 32 dimensions, the PIM system
takes about 73 seconds to compute ${32 \choose 3}=4,960$ aggregates, demonstrating scalability with the number
of aggregates computed.

Figure~\ref{fig:col_test_taxi} shows that for datasets with 16 or more dimensions, using PIM in asynchronous mode results in 45\% to 64\%, 28\% to 30\%, and 31\% to 53\% speedup compared to the CPU machine for integer, float and double data type aggregates, respectively. The speedup for float data types is lower than for integer data types as doing floating point calculations on DPUs is significantly slower than doing integer calculations, whereas on CPUs, it is only slightly slower. For double data types, the speedup from using PIM is slightly higher than floats; while double data type calculations in the DPU do take longer compared to floats, since the DPU is only able to write to memory at a minimum of 8 bytes at a time, there is no slowdown regarding memory bandwidth, whereas the CPU machine needs to write to twice the memory. 

For datasets with 8 dimensions, using PIM in asynchronous mode is 25\% to 41\% slower compared to the CPU machine. On datasets with comparatively few dimensions, there are fewer aggregates to update, so relatively more of the PIM execution time is spent filling buffers in the host CPU and transferring data to the DPUs compared to updating aggregates in the DPUs, as shown in Figure~\ref{fig:col_test_taxi} using PIM in synchronous mode. As the PIM server uses a relatively weak CPU that was released in 2019, it would be possible to achieve better performance for the PIM server on datasets with fewer dimensions by replacing the PIM server's CPU with a more modern CPU.

Figure~\ref{fig:col_test_taxi} shows that using PIM in asynchronous mode has the intended effect of overlapping time spent filling buffers in the host CPU and updating aggregates in the DPUs.  End to end execution time is reduced relative to synchronous execution by the smaller of the time spent updating aggregates or filling host CPU buffers.

Execution times for the flight dataset are shown in Figure~\ref{fig:col_test_flight}. Since time spent updating aggregates, filling buffers in the host CPU, and transferring data from the host CPU to DPUs is linear to the number of rows in the dataset, for the PIM system we get speedups similar to those shown in the taxi dataset against the CPU machine to those sections of execution. However, since the total size of the aggregations depends solely on the number of dimensions, for the same number of dimensions the time to transfer the aggregates back to the host CPU is the same as those shown in the taxi dataset, meaning this transfer time takes up a larger portion of execution time. Overall, using PIM in asynchronous mode results in a 25\% to 49\% speedup compared to the CPU machine on 16 or more dimensions, and is 44\% to 69\% slower than the CPU machine for 8 dimensions, which could be improved by replacing the PIM system's CPU with a faster CPU having more memory bandwidth.

Although the hardware on the PIM system was released in 2019, we believe it is fairest to compare our PIM system with a modern CPU. We have also compared the PIM system with a Intel Xeon Gold 6312U CPU released in 2021, and observed approximately 
6X speedup when using the PIM system on the taxi dataset. A hypothetical PIM system using modern (2024) technology could potentially achieve similar speedups compared to modern CPUs.

\subsection{Varying the Number of Bins}

We compare runtimes of calculating aggregates on the taxi dataset that is binned with either 32, 64 or 128 bins. Results are shown in Figure \ref{fig:bin_test} for 24 dimensions. 
\eat{We show results for calculating integer aggregates for 24 dimensions, and omit results for other number of dimensions and data types as the trends were similar.
}
For the general purpose CPU, we can see that execution times for 32 and 64 bins is approximately 11X and 6X faster, respectively, than for 128 bins. This is because when we use 64 or fewer bins, a size of a 3-dimensional aggregate is 1MiB or less, which fits inside the CPU's 2MiB L2 cache.
For PIM, we only report the end-to-end execution time for 128 bins as the end-to-end execution time was similar for all bin sizes. This is because each DPU receives the same number of tuples, and writes to the same number of locations in memory regardless of the bin size. While a smaller bin size means that a subgroup would receive more tuples, our optimization of duplicating aggregates and partitioning tuples among DPUs when there are leftover DPUs ensures that each DPU receives the same number of tuples for smaller bin sizes.

For our largest experiments with 32 dimensions and 128 bins, the system needs to store 10.4 billion data elements, or about 41GB using four-byte aggregates. 256 bins would not be feasible on our PIM system as it would require more than the 160GB available. To improve the resolution by using 256 bins, one would need to limit the number of dimensions to 24. Even in such cases, the 
cost of transferring the results from the DPU to the CPU increase by a factor of 8, and become a dominant factor in the overall cost. Thus, for the
DPU system, a bin size of 128 is a sweet-spot that balances performance with visualization image resolution.

\subsection{Varying the Number of DPUs}

\begin{figure}
    \centering
    \includegraphics[width=0.7\columnwidth]{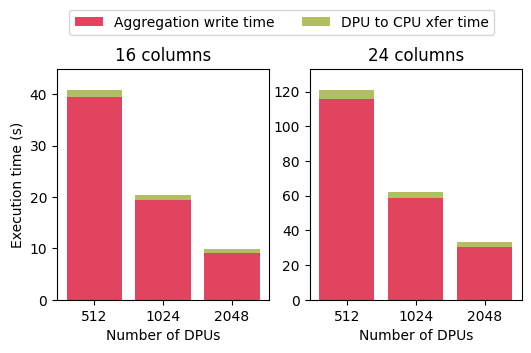}
    \vspace{-1mm}
    \caption{Execution time with various numbers of DPUs}
    \label{fig:dpu-num-test}
\end{figure}

We measure the end to end execution time of calculating integer aggregates on the taxi dataset with 16 and 24 dimensions, with varying number of DPUs running in asynchronous mode. Results are shown in Figure ~\ref{fig:dpu-num-test}. We can see that increasing the number of DPUs results in speedup almost proportional to the number of DPUs. This shows that our aggregation distribution approach is flexible and scales well with the number of DPUs. Also, while not proportional to the number of DPUs, using more DPUs results in less time spent transferring aggregates to the host CPU as it is more efficient to transfer the same amount of data using more DPUs. \eat{Overall, we can see that a PIM system can calculate aggregates faster on large datasets by simply installing more DPUs, as opposed to introducing new hardware.}

\subsection{CPU Experiments}
\label{sect-exp-CPU}

\begin{figure}
    \centering
    \includegraphics[scale=0.5]{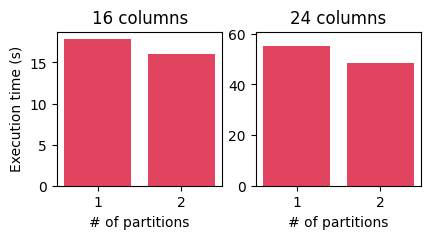}
    \vspace{-1mm}
    \caption{Execution time where aggregate is partitioned}
    \label{fig:partition-test}
\end{figure}

We ran experiments to see the effect of dividing aggregate arrays into smaller partitions, so that we can get better temporal locality at the cost of scanning more data. Results are shown in Figure \ref{fig:partition-test} for calculating integer aggregates on the taxi dataset for 16 and 24 dimensions. \eat{We omit results for other data types as the speed ups were similar, and} Dividing the aggregate into 4 partitions resulted in no speedup. We can see that dividing the aggregate into 2 partitions results in approximately 13\% speedup. As the size of each 3 dimensional aggregate is 8MiB, and the size of the L2 cache on the CPU machine for each core is 2MiB, by dividing the aggregate into two partitions we reduce the number L2 cache misses. While dividing the aggregate into 4 partitions would make it fit in the L2 cache, the system would need 2 more passes over the whole dataset, which is why we do not get additional speed ups.

\subsection{Visualization Comparison}\label{s:vis_comp}

\begin{figure*}
    \centering
    \begin{minipage}{0.50\columnwidth}
        \centering
        \includegraphics[width=1.0\textwidth]{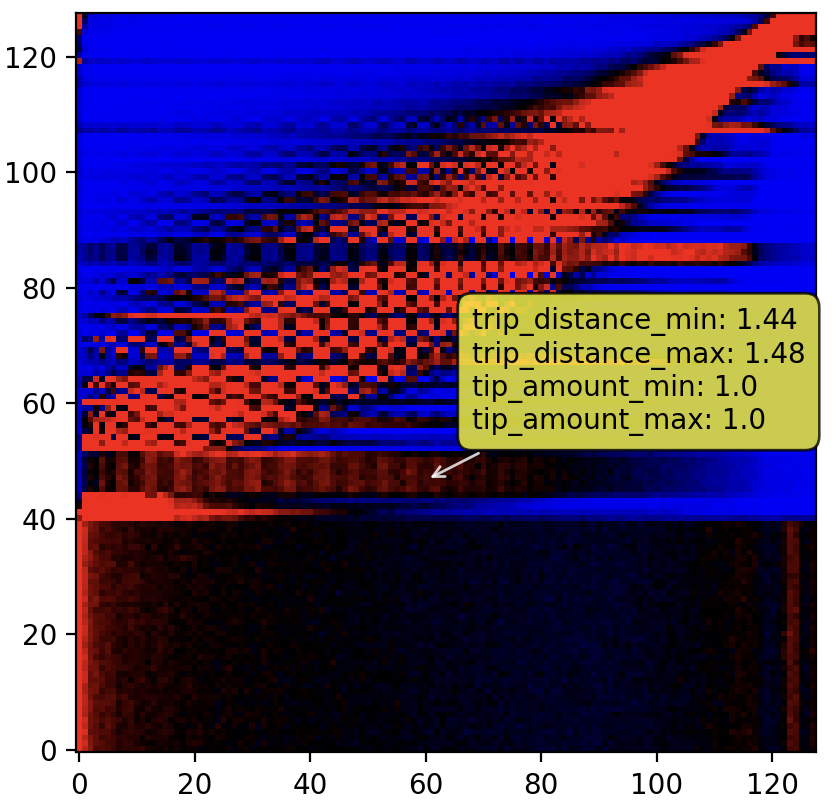}
    \\ (a)
    \end{minipage}\hfill
    \begin{minipage}{0.50\columnwidth}
        \centering
        \includegraphics[width=1.0\textwidth]{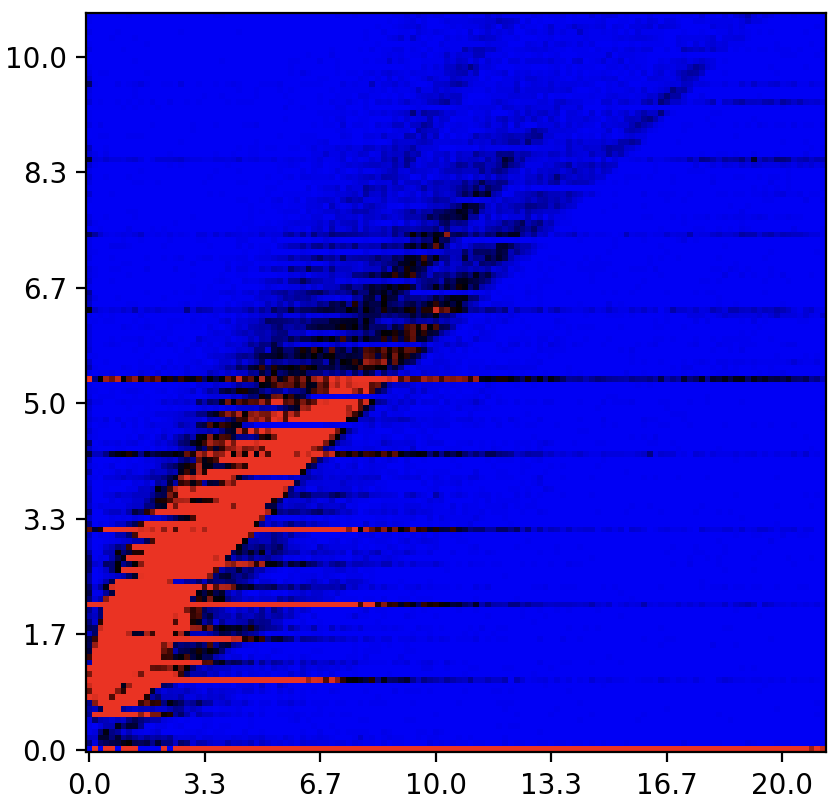} 
        \\ (b)
    \end{minipage}\hfill
    \begin{minipage}{0.50\columnwidth}
        \centering
        \includegraphics[width=1.0\textwidth]{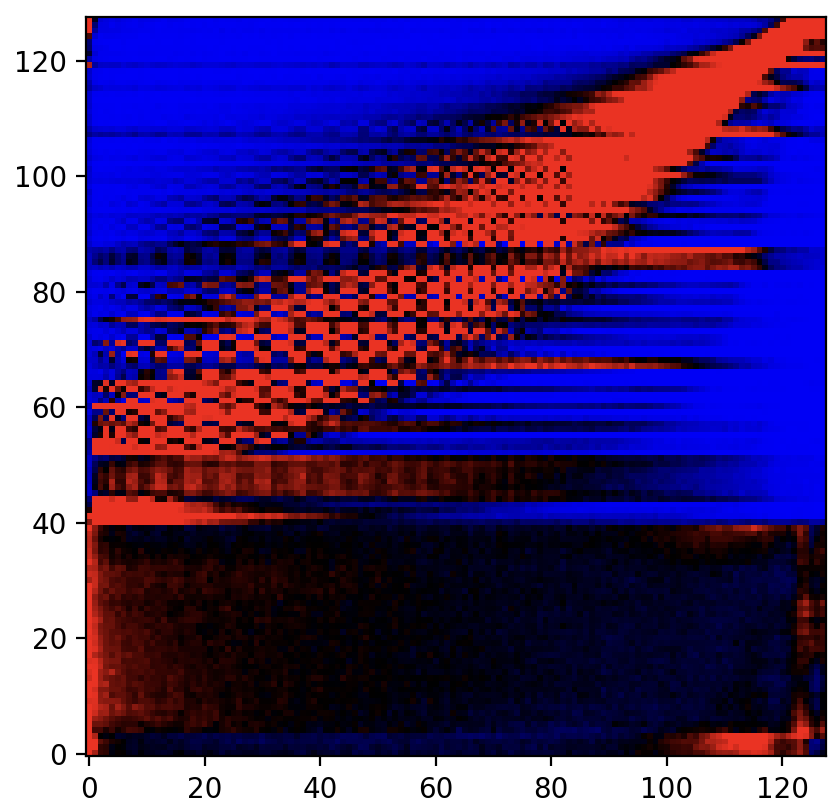} 
        \\ (c)
    \end{minipage}
    \begin{minipage}{0.50\columnwidth}
        \centering
        \includegraphics[width=1.0\textwidth]{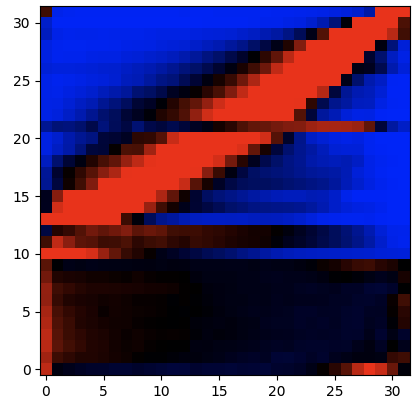} 
        \\ (d)
    \end{minipage}
    \vspace{-1mm}
    \caption{Visualization between tip amount (y-axis) and trip distance (x-axis), using \vispim{}'s binning (a) and equidistance binning (b) using 128 bins. In diagrams (c) and (d), the y-axis is 
    (tip-amount,pickup-time), binned with 128 and 32 bins respectively.}
    \label{fig:tip-trip-cmp}
\end{figure*}

\begin{figure}
    \centering
    \begin{minipage}{0.5\columnwidth}
        \centering
        \includegraphics[width=0.95\textwidth]{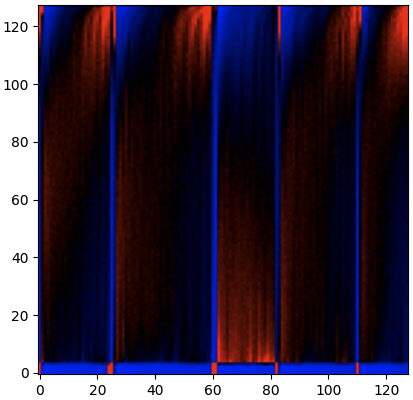}
    \\ (a)
    \end{minipage}\hfill
    \begin{minipage}{0.5\columnwidth}
        \centering
        \includegraphics[width=1.0\textwidth]{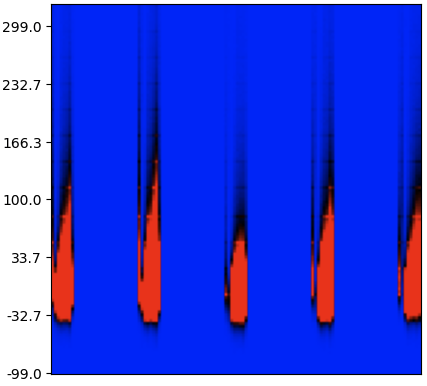} 
        \\ (b)
    \end{minipage}\hfill
    \vspace{-1mm}
    \caption{Visualization between arrival delay time (y-axis) and 
    (year, arrival time) (x-axis), using \vispim{}'s binning (a) and equidistance binning (b) with 128 bins.}
    \label{fig:year-arrdelay-cmp}
\end{figure}

We present images generated by our system for the taxi dataset and flight dataset.
For analysis, the user can mouse over the image to get the range of values corresponding to the bin at the location of the cursor, as shown in Figure~\ref{fig:tip-trip-cmp}(a).
We also generate images using a conventional equidistance binning of the original dimension domains. 
For an apples-to-apples comparison, we implemented a pixel computation function for equidistance binning that highlights
cells based on over-representation or under-representation relative to the expected value according to the marginal
distributions of each dimension. 

\eat{
\reminder{Remove this, and replace with images using tic marks}
For simplicity, we do not show tic marks with domain values on the visualization itself. In practice, this information
is easy to obtain by interacting with the visualization, e.g., by mousing over a coordinate and having a 
tooltip appear with the $x$ and $y$ ranges (in the original dimension scales) for the current cell in the image.
}


We begin by comparing two visualizations generated using count aggregates between the tip amount and trip distance dimensions in the taxi dataset. For the first visualization (Figure~\ref{fig:tip-trip-cmp}(a)), we use \vispim{} to calculate aggregates, using 128 bins for equidepth binning. For the second visualization (Figure~\ref{fig:tip-trip-cmp}(b)), we use equidistance binning with 128 bins within the tip and distance domains.

The robustness of our visualization approach is illustrated by the steps necessary to generate these images. Our initial attempts to generate an equidistance visualization were confounded
by some erroneous values, including over 900 negative values for the tip and one tip amount of over \$141,000! Such values would
distort the equidistance diagram to the point of being useless. Thus, we were forced to apply a manual data cleaning step in which
we identified reasonable bounds for the tip and distance, and filtered the dataset to remove outliers. Figure ~\ref{fig:tip-trip-cmp} (b) corresponds to the subset of the data with tip amounts between \$0 and \$15 and trip distances between 0 and 20 miles.
Even with these reasonable limits, Figure~\ref{fig:tip-trip-cmp}(b) uses a lot of visualization real-estate for regions with
little data, colored blue. In contrast, our equidepth binning is robust to a small proportion of outliers and
uses the original full data set data directly. Further, Figure~\ref{fig:tip-trip-cmp}(a) uses image real-estate in proportion
to data populations.

Even before visualizing the data, one expects to see a positive correlation between the trip distance and the tip amount.
Our visualization allows for a deeper understanding of this correlation.
Firstly, there is a large black rectangle at the bottom of Figure~\ref{fig:tip-trip-cmp}(a) that spans 39  bins vertically and 128 bins horizontally. This region corresponds to tip amounts that are \$0, and we can see that approximately 30\% of riders give no tip, and that giving no tip is common regardless of the trip distance. In the equidistance representation of Figure~\ref{fig:tip-trip-cmp}(b), this pattern corresponds to the thin red line at the bottom of the visualization. This line is not immediately obvious, and it is hard to gauge what percentage of riders do not tip.

Secondly, in Figure~\ref{fig:tip-trip-cmp}(a) horizontal lines span several bins, interrupting the correlation. Upon inspection by mousing over these lines as shown in the figure, these values correspond to rounded tip values, such as \$1, \$3 or \$5, meaning these tip values are more common compared to other positive tip values, and users are more inclined to give these values as tip, even if it is not at the normal rate for the trip distance. For Figure~\ref{fig:tip-trip-cmp}(b), there are horizontal lines sticking out of the correlation,
but it is far from clear that there is a horizontal displacement in the distribution for discrete dollar tips rather than just a wider distribution.

\begin{figure*}
    \centering
    \begin{minipage}{0.52\columnwidth}
        \centering
        \includegraphics[width=0.95\textwidth]{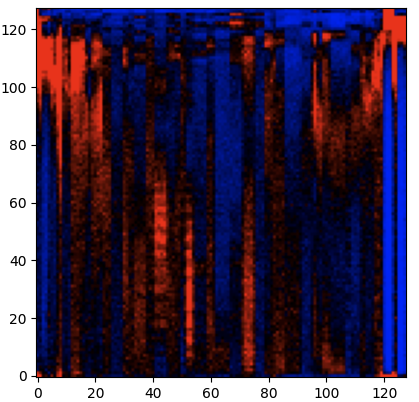}
    \end{minipage}\hfill
    \begin{minipage}{0.52\columnwidth}
        \centering
        \includegraphics[width=0.95\textwidth]{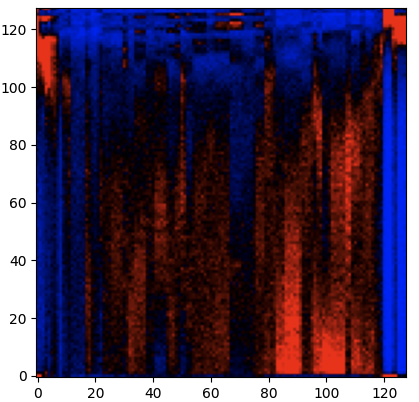}
    \end{minipage}\hfill
    \begin{minipage}{0.52\columnwidth}
        \centering
        \includegraphics[width=0.95\textwidth]{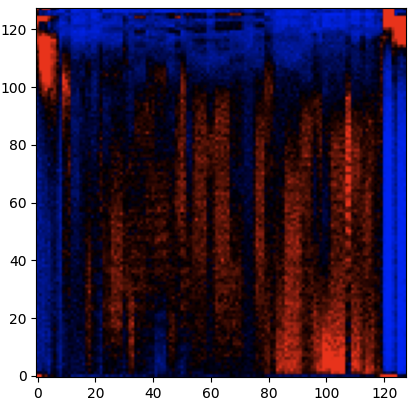}
    \end{minipage}\hfill
    \begin{minipage}{0.52\columnwidth}
        \centering
        \includegraphics[width=1.10\textwidth]{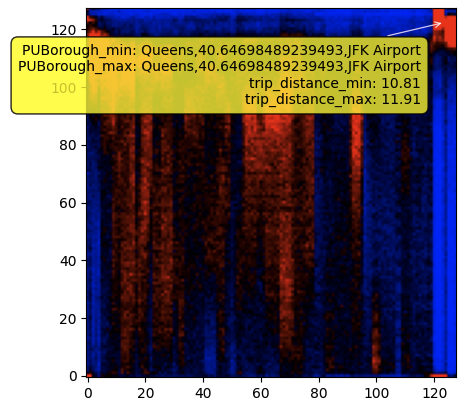}
    \end{minipage}
    \caption{Visualizations created using \vispim{} of aggregates between pickup location (x-axis) and trip distance (y-axis), partitioned into 4 groups using pickup time (z-axis)}
    \label{fig:tip-trip-cmp2}
\end{figure*}

\begin{figure*}
    \centering
    \begin{minipage}{0.52\columnwidth}
        \centering
        \includegraphics[width=1.0\textwidth]{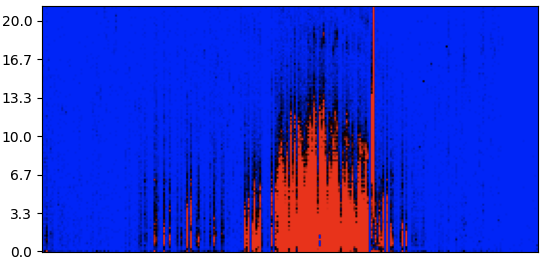}
    \end{minipage}\hfill
    \begin{minipage}{0.52\columnwidth}
        \centering
        \includegraphics[width=1.0\textwidth]{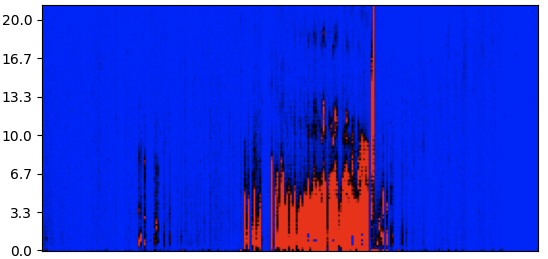}
    \end{minipage}\hfill
    \begin{minipage}{0.52\columnwidth}
        \centering
        \includegraphics[width=1.0\textwidth]{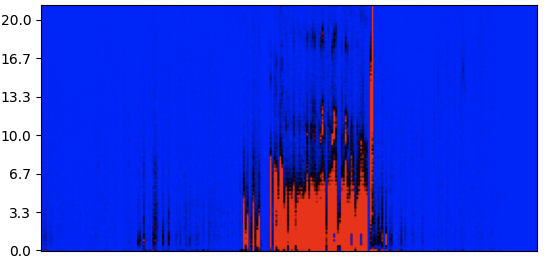}
    \end{minipage}\hfill
    \begin{minipage}{0.52\columnwidth}
        \centering
        \includegraphics[width=1.0\textwidth]{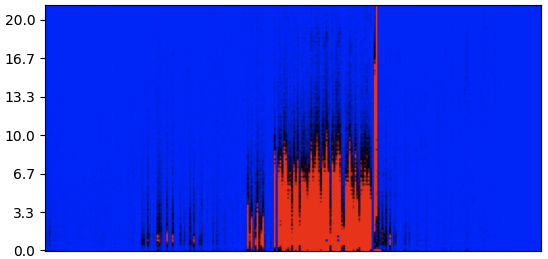}
    \end{minipage}
    \caption{Visualizations of aggregates calculated on a categorical/equiwidth binned dataset, between dimensions pickup location (x-axis) and trip distance (y-axis), partitioned into 4 groups using pickup time (z-axis)}
    \label{fig:tip-trip-cmp-width}
\end{figure*}

Figure~\ref{fig:tip-trip-cmp}(a) does waste some visualization capacity, because all of the 39 vertical bins at the bottom of the diagram
have the same horizontal distribution. One way to improve the visualization is to use a secondary attribute. Figure~\ref{fig:tip-trip-cmp}(c)
is similar to Figure~\ref{fig:tip-trip-cmp}(a) except that the $y$ dimension is now the pair (tip-amount,pickup-time).
As a result, ties in tip-amount will be ordered by pickup-time. The bottom 39 bins now
reveal a new pattern, namely that there is an over-representation of \$0 tips for long trips early in the day. Figure \ref{fig:tip-trip-cmp}(d) uses the same dimensions as Figure \ref{fig:tip-trip-cmp}(c), except that the data is binned using 32 bins. With a lower resolution, it becomes harder to do fine-grained analysis, as some horizontal lines corresponding to discrete tip values are no longer apparent.

We also compare two visualizations generated using count aggregates between the arrival delay and the pair (year, arrival time) dimensions in the flight dataset, shown in Figure \ref{fig:year-arrdelay-cmp}. For the visualization generated by \vispim{}, the bins for each year is further divided by arrival time, which shows additional patterns for each year. All years have a general pattern where flights with arrival times late in the night also tend to have arrival delays. However, we can easily notice that the middle year, which corresponds to 2020, has a less strong correlation compared to other years. The bright dots at the bottom of each year are null values for arrival delays that take up around 3\% of the data. \eat{This shows that \vispim{} generates robust visualizations that allocate less space for rare outliers.} For Figure \ref{fig:year-arrdelay-cmp}(b), we tried to create a comparable visualization by concatenating the year and arrival time into a single integer, and then applying equidistance binning. While we do get separate patterns for each year, we get skewed bins due to the magnitude of year, and the positive correlation between arrival time and arrival delay is harder to see.


We next show an example of image groups partitioned by a $z$ attribute.
We consider pickup location as the $x$ dimension, trip distance as the
$y$ dimension, and pickup time as the $z$ dimension.
In the taxi dataset, pickup and dropoff locations are coded into 265
discrete zones.
For the $x$ dimension we use a composite (borough,latitude) representation so that zones are
ordered primarily by borough (e.g., Manhattan, Queens), and south-to-north within each borough according to the latitude of the
zone's centroid.
Figure~\ref{fig:tip-trip-cmp2} shows our generated visualizations for four $z$ partitions.
Figure~\ref{fig:tip-trip-cmp-width} shows a version of the same data in which each unique pickup
location has its own bin; the x-axis is longer (265 bins) compared to the equidistance-binned
y-axis (128 bins).

The vast majority of the data corresponds to pickups in Manhattan. Correspondingly, the
majority of the horizontal axis in Figure~\ref{fig:tip-trip-cmp2} (the central region) is devoted to
Manhattan, unlike Figure~\ref{fig:tip-trip-cmp-width}.
The cluster at the top-right of the images in Figure~\ref{fig:tip-trip-cmp2}, highlighted using the interactive tool in Figure~\ref{fig:tip-trip-cmp2}(d), corresponds to long trips
from the airports JFK and LGA, which are located in Queens. In Figure~\ref{fig:tip-trip-cmp-width}, the airports correspond to the tall vertical line that is at the right end of the cluster.

Figure~\ref{fig:tip-trip-cmp2} shows that the location/distance interaction changes depending on the time, something that
is much harder to appreciate from Figure~\ref{fig:tip-trip-cmp-width}. We can see that within Manhattan, which is geographically vertically long, people tend to start long taxi rides in the lower part of Manhattan at the beginning and end of day, and take shorter taxi rides in the higher part of Manhattan later in the middle of day, as inspecting the data showed that the middle 2 images correspond to pickup times between 10am and 6pm.

\subsection{Visualization Recommendation Evaluation}

\begin{figure}
    \centering
    \includegraphics[width=\columnwidth]{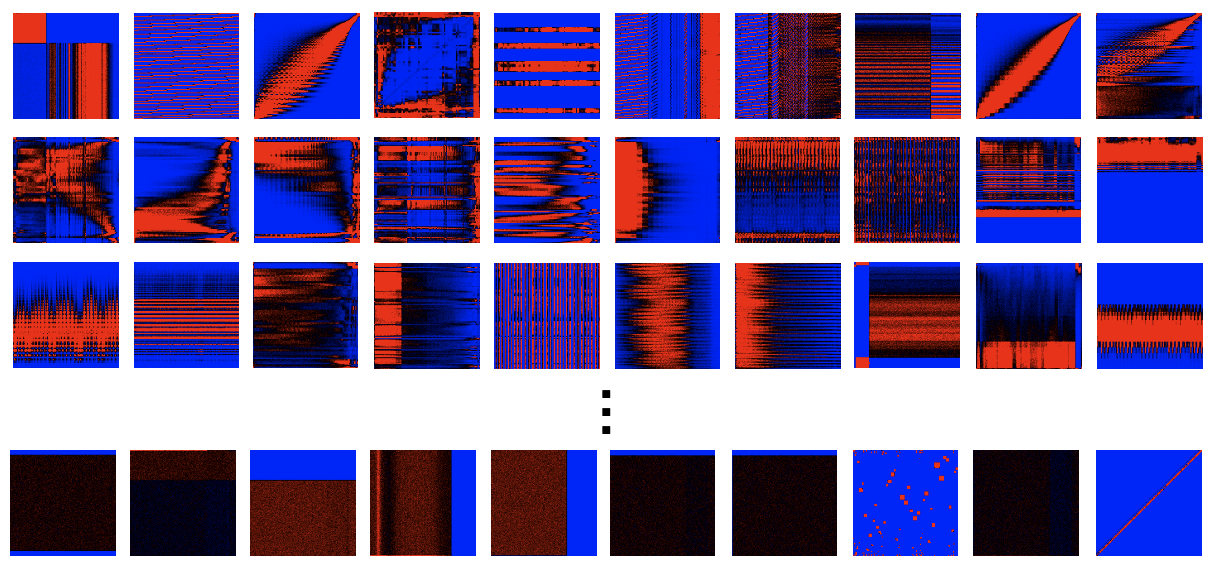}
    \caption{Images from top 30 and bottom 10 groups}
    \label{fig:ordering}
\end{figure}

We evaluate \vispim{}'s ability to present interesting images to the user using the process described in Section \ref{s:vis recommendation}. For this experiment, we use the 16-dimension taxi dataset and compute count aggregates, and generate 12 images for every 3-dimensional aggregate by dividing each dimension into 4 partitions, for a total of 6720 images. Figure \ref{fig:ordering} shows one image from each of the top 30 groups, along with the highest scoring images of the bottom 10 groups. This heuristic successfully produces images that have distinct, interpretable patterns, including visualizations shown in Section~\ref{s:vis_comp}.\eat{Implementing a more sophisticated visualization recommendation system is beyond the scope of this paper, and we leave this problem to future work.}

\section{Related Work}

Data cubes~\cite{datacubes} are a common candidate for visualization for the purpose of identifying correlations. \cite{hashedcubes, nanocubes, liu2013immens} propose data structures that require less memory than a traditional data cube by computing aggregates on binned data, with the purpose of supporting interactive querying times on the aggregates and saving storage. However, they do not consider the normalized binning method introduced in this paper. \cite{maroulis2023resource, maroulis2021rawvis} explores how adaptive indexing can be used to efficiently support data visualization tasks on commodity hardware on large data files that do not fit in memory. \cite{wu2024multirelationalalgebraapplicationsdata} proposes an extension of relational algebra for the purpose of expressing complex analytic queries on aggregates. The goal of \vispim{} is related to data profiling~\cite{profilingnaumann}, the process of collecting information and statistics on data. \cite{sarikaya2017scatterplots} discusses several ways to visualize scatterplots for large amounts of data, depending on the data's characteristics. \cite{battlestructreview} provides a comprehensive review of recent trends at the intersection of visualization and databases.

Our pixel computation function is related to the usage of heatmaps in genomic data analysis, where over/under-representation has biological significance~\cite{Gu2016ComplexHR}.

With the release of commerical hardware that supports Processing-in-Memory there has been work on utilizing it in the field of databases, in particular for accelerating joins \cite{dimmjoin}, ordered indexes \cite{pimtree}, large table scans \cite{baumstark2023accelerating}, adaptive query processing \cite{baumstark2023adaptive} and integration of PIM in main-memory DBMSs \cite{pimdb}.

\eat{
\reminder{Don't forget to cite the review articles from Eugene's email and reference them as comprehensive reviews of visualization in data management.}
}

Our preprocessing step is related to data discretization~\cite{ramirez2016data} in that we convert the domain of the database's dimensions to integers which are easier to work on. Our approximate binning method is related to approximating frequencies for building equidepth histograms~\cite{poosala1996improved} but differs in how the bin boundaries are handled.

\eat{
Our workflow is similar to that of a visualization recommendation system~\cite{vartak2017towards, vizml}, a system that helps fast analysis of data by automatically recommending visualizations the user might be interested in.
\reminder{But how is our work different?}
}

\section{Conclusions}\label{s:conclusions}

We propose \vispim{}, a system that utilizes normalized binning to create visualizations of aggregates that automatically focuses on where the data is. \vispim{} is able to quickly calculate all possible 3-dimensional aggregates of the binned dataset using either general purpose CPUs or Processing-in-Memory architectures, using algorithms that take advantage of each hardware's memory characteristics. By using Processing-in-Memory, \vispim{} is able to calculate aggregates on large scale datasets that have tens of millions of rows and tens of columns 45\%-64\% faster than modern CPUs. \vispim{} is fully end-to-end in that it accepts a dataset, and automatically bins the dataset and calculates the aggregates using the available hardware it is on, and recommends interesting visualizations for the user to analyze. We show through examples that \vispim{} generates visualizations that highlight both expected and unexpected patterns in the data that are more complex than simple correlations.

For future work, we plan to investigate more effective ways of doing visualization recommendation~\cite{vartak2017towards, vizml} by automatically identifying interesting patterns in the images generated by \vispim{} using more sophisticated techniques. 
\eat{, and extend \vispim{} to support a more interactive workflow.}

\begin{acks}
This work was supported by the National Science Foundation under grant III-2312991.
\end{acks}

\bibliographystyle{ACM-Reference-Format}
\bibliography{references}

\end{document}